\documentclass[10pt,twocolumn,letterpaper]{article}
\pdfoutput=1
\usepackage{cvpr}
\usepackage{times}
\usepackage{epsfig}
\usepackage{graphicx}
\usepackage{amsmath}
\usepackage{amssymb}
\usepackage{booktabs}
\usepackage{caption}
\usepackage{float} 
\usepackage{subcaption}
\usepackage{multirow}
\usepackage[table,xcdraw]{xcolor}

\usepackage[pagebackref=true,breaklinks=true,letterpaper=true,colorlinks,bookmarks=false]{hyperref}

\newcommand\blfootnote[1]{%
\begingroup
\renewcommand\thefootnote{}\footnote{#1}%
\addtocounter{footnote}{-1}%
\endgroup
}

\cvprfinalcopy % *** Uncomment this line for the final submission

\def\cvprPaperID{13} % *** Enter the WSS Paper ID here

\ifcvprfinal\fi
\begin{document}

\title{Balanced Audiovisual Dataset for Imbalance Analysis}  % **** Enter the paper title here

\author{
Wenke Xia\textsuperscript{1,$\dagger$}, 
Xu Zhao\textsuperscript{2,$\dagger$}, 
Xincheng Pang\textsuperscript{1}, 
Changqing Zhang\textsuperscript{3}, 
Di Hu\textsuperscript{1,4,$\ast$}
\vspace{0.5em}
\\
\textsuperscript{1}Gaoling School of Artificial Intelligence, Renmin University of China, Beijing\\
\textsuperscript{2}School of Computer Science and Engineering, Northeastern University, Shenyang \\
\textsuperscript{3}College of Intelligence and Computing, Tianjin University, Tianjin \\
\textsuperscript{4}Beijing Key Laboratory of Big Data Management and Analysis Methods, Beijing\\
Project website: \href{https://gewu-lab.github.io/Balanced-Audiovisual-Dataset/}{https://gewu-lab.github.io/Balanced-Audiovisual-Dataset/}
}

\maketitle
\thispagestyle{empty}
\blfootnote{$\dagger$ Equal contribution. $\ast$ Corresponding author.}

\section{Introduction}

\begin{figure*}[h]
    \centering
		\centering
		\includegraphics[width=\textwidth]{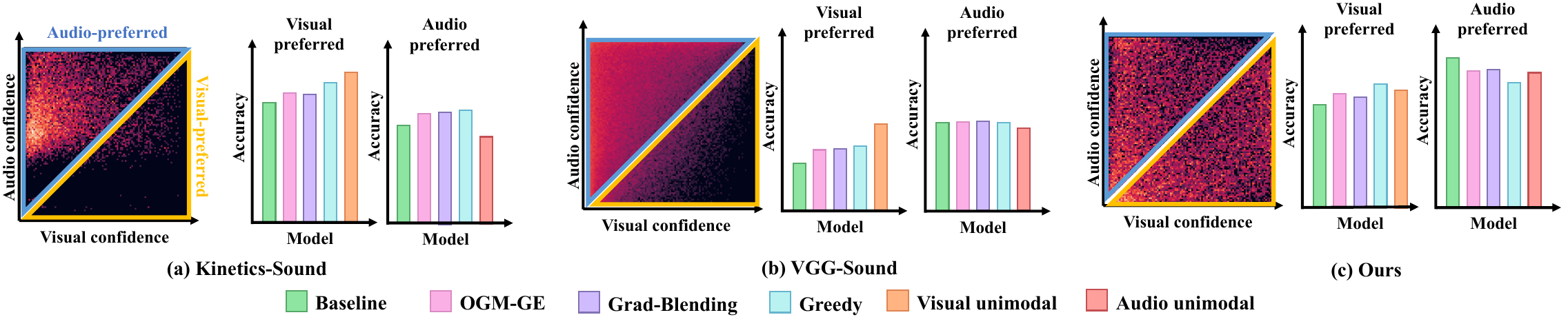}
    \caption{(\textbf{a}-\textbf{b}) show the heatmaps of modality discrepancy distribution and classification results on the existing datasets. In the heatmap, the color represents the proportion of samples, the brighter area indicates larger samples. (\textbf{c}) shows our balanced heatmap and the classification results.}
    \label{fig:heatmaps}
    \vspace{-1em}
\end{figure*}

The issue of imbalance has garnered significant attention across various domains, including loss imbalance in multi-task learning, category imbalance in long-tailed datasets, etc. Imbalance problems are also prevalent in the area of multimodal learning, where models tend to rely on the dominant modality in the presence of discrepancies between modalities~\cite{peng2022balanced}.
Recent studies~\cite{peng2022balanced,wang2020makes} have endeavored to address the issue of modality imbalance. Wang et al.~\cite{wang2020makes} found the phenomenon that the performance of multimodal models would be inferior to the unimodal model due to the modality overfitting behaviors. Peng et al.~\cite{peng2022balanced} and Wu et al.~\cite{wu2022characterizing} proposed to solve the modality imbalance problem with the perspective of the optimization process, while Han et al.~\cite{han2022trusted} alleviated this  issue by making reliable multi-view fusion. However, we observe that although existing imbalance methods exhibit superior performance on the overall testing set, they even fail to perform better than the unimodal model when inference on some modality-preferred subsets as illustrated in Figure~\ref{fig:heatmaps}. This phenomenon has not been adequately addressed in prior works but it may impact the reliability of multimodal models in certain scenarios, such as those involving modality noise.

To provide a comprehensive analysis of this phenomenon,  we first introduce a metric to estimate the sample-wise modality discrepancy with unimodal confidence and divide existing datasets into different modality-preferred subsets. Concretely, we define a modality-preferred subset as a set of samples whose corresponding modality confidence is higher than the other modality. Further, we evaluate existing imbalance methods and find that: although existing imbalance methods achieve better performance than unimodal modality on the overall testing set, they consistently perform worse than the unimodal model on the visual preferred testing subset across different datasets as demonstrated in Figure~\ref{fig:heatmaps}(a-b).

However, given the data-driven nature of deep learning~\cite{wu2022characterizing}, it is inevitable that the serious modality bias present in existing datasets such as Kinetics-Sound~\cite{arandjelovic2017KS} and VGG-Sound~\cite{chen2020vggsound} will cause multimodal models to learn this bias.
Thus, to further investigate the effectiveness of imbalance methods, we build a balanced audiovisual dataset containing samples with various modality discrepancies, and the allocation of such modality discrepancy remains uniformly distributed over the dataset, as shown in Figure~\ref{fig:heatmaps}(c). 

Based on our balanced dataset, we re-evaluate existing imbalance methods and find results consistent with the previous phenomenon: there still exists no imbalance method exceeding all unimodal models on every modality-preferred testing subset as shown in the histogram of Figure~\ref{fig:heatmaps}(c).
We further select testing subsets with different modality discrepancies to evaluate the performance of existing imbalance methods and surprisingly find that the larger the modality discrepancy is, the worse the multimodal models perform compared with the unimodal one. This phenomenon indicates that existing imbalance methods have difficulty handling multimodal data with large modality discrepancy. Thus, the evaluation of different modalities discrepancies should also be considered to comprehensively analyze the performance of multimodal models.

To summarize, our contributions are as follows:
\begin{itemize}
    \item We introduce an evaluation metric to estimate the modality discrepancy and point out that existing audiovisual datasets suffer from serious modality bias.
    \item We provide a balanced audiovisual dataset with uniformly distributed modality discrepancies to support further analysis for multimodal models.
    \item We find that existing multimodal models would exhibit inferior performance than the unimodal one in scenarios with serious modality discrepancy. We highlight the importance of addressing this issue in future research to ensure the reliability of multimodal models across various application scenarios.
\end{itemize}
% Besides, when increasing the degree of modality discrepancy, the performance is worse.
% This phenomenon indicates that existing imbalance methods still xx, the evaluation in different modalities discrepancies should be considered to make sure the reliability of multimodal models.

% Thus, we propose an evaluation method to estimate the degree of modality discrepancy, and further evaluate the performance of imbalance methods in scenarios with different degrees of modality imbalance. We also provide a balanced audio-visual dataset to promote further analysis of multimodal models.

\section{Evaluation Methods}

\subsection{Modality discrepancy estimation}

Inspired by some sample difficulty measure work in unimodal datasets~\cite{ethayarajh2021pvi}, we provide an estimation method for measuring the sample-wise modality discrepancy, which utilizes the difference between unimodal confidence assigning to the true label during the training process. 

Concretely, we consider a training dataset of size $N$, $S = \{{(X_i, Y_i)}\}_{i=1}^N$ where the $i$th sample contains the observation, $X_i$ and its true label, $Y_i$. In multimodal datasets, we consider each observation $X_i$ consists of $M$ inputs from different modalities, referred as $X_i = (x^{m_1}_i,..,x^{m_k}_i,..,x^{m_M}_i)$, where $m_k$ refer to the $k$th modality respectively. We first train each unimodal model with an individual modality for $E_{m_k}$ epochs, and utilize the average confidence of unimodal to represent the learning difficulty for the $i$th sample:
\begin{center}
$\hat u_i^{m_k} = \frac{1}{E_{m_k}} \sum\limits_{e=1}^{E_{m_k}}p^{m_k}_{\theta}(e)(y_i|x_i^{m_k}) $,
\end{center}
where $p^{m_k}_{\theta}(e)$ presents the confidence of unimodal model with parameters $\theta$ at the $e$th epoch.

We divide such average unimodal confidence scores into 0.01 intervals, then count the samples of corresponding intervals and calculate the proportion, which could be regarded as a distribution of modality discrepancy. 
We present the resulting distribution as a heatmap in Figure~\ref{fig:heatmaps}, where the coordinates correspond to the average unimodal confidence scores and the color represents the proportion of samples within the corresponding interval. Our method thus provides an intuitive method to visualize the modality discrepancies in multimodal datasets.

\subsection{Dataset split strategy}
We use the average unimodal confidence $\hat u_i^a$ and $\hat{u_i^v}$, to represent the unimodal confidence on audio and visual modalities. To investigate modality bias in audiovisual datasets, we split an existing audiovisual dataset into two subsets based on the unimodal confidence scores. We name these subsets the audio-preferred subset and the visual-preferred subset. If $\hat{u_i^a} > \hat{u_i^v}$, the sample would be split into the audio-preferred subset, while if $\hat{u_i^v} > \hat{u_i^a}$, the sample would be split into the visual-preferred subset instead.

With our split strategy, we divide the testing set of existing datasets into different modality-preferred subsets. As depicted in Figure~\ref{fig:heatmaps}(a-b), our estimation results prove that there exists serious modality bias in these audiovisual datasets, which may not be adequate for comprehensive analysis of the existing modality imbalance methods.

% As depicted in Figure~\ref{fig:heatmaps}(a-b), our analysis reveals that the audio-preferred subset is much larger than the visual-preferred subset in the Kinetics-Sound and VGG-Sound datasets. This modality discrepancy estimation results prove that there exists serious modality bias in these audiovisual datasets, which may not be adequate for comprehensive analysis of the existing modality imbalance methods.

\begin{table*}[h]
\centering
\resizebox{0.8\textwidth}{!}{
\begin{tabular}{c|ccc|ccc}
\toprule
                & \multicolumn{3}{c|}{Kinetics-Sound}      & \multicolumn{3}{c}{VGG-Sound}            \\ \midrule
Methods         & Overall   & Audio preferred & Visual preferred & Overall   & Audio preferred & Visual preferred \\ \midrule
Baseline        & 61.73 & 60.96          & 69.09           & 48.65 & 52.38          & 27.50           \\
OGM-GE          & 64.57 & 63.20          & 74.44           & 50.89 & 53.43          & 35.63          \\
Grad-Blending   & 66.00 & 64.34          & 73.82           & \textbf{51.47} & \textbf{53.73}        & 35.87           \\
Greedy          & \textbf{66.35} & \textbf{64.51}          & 80.75           & 50.32 & 52.68         & 37.37         \\ \midrule
Audio unimodal  & 51.52 & 54.32          & 31.23           & 41.25 & 45.39          & 10.86          \\
Visual unimodal & 47.67 & 42.28          & \textbf{86.43}           & 32.44 & 28.65          & \textbf{51.67}           \\
\bottomrule
\end{tabular}}
    \caption{Evaluation results on the existing audiovisual dataset.}
    \label{tab:av_other}
    \vspace{-1em}
\end{table*}

\section{Experiments on Existing Datasets}
\subsection{Experiments settings}

We estimate the modality discrepancy of Kinetics-Sound~\cite{kay2017kinetics} and VGG-Sound~\cite{chen2020vggsound} datasets, and further, evaluate existing imbalance methods on these datasets. We use ResNet-18 as the backbone for both modalities. Following Peng et al.~\cite{peng2022balanced}, for audio input, we extract the spectrogram with a sample rate of 48$kHz$, while we extract frames with 1fps and randomly select frames as the visual input. 

For modality discrepancy estimation, we train each modality separately to evaluate the unimodal confidence. Given that the coverage speed of each modality varies in different datasets, we adopt an early-stop strategy during training to obtain reliable unimodal confidence.

\subsection{Experiments results}
In this section, we conduct a comprehensive evaluation of existing imbalance methods using different subsets, with a vanilla late-fusion method as the baseline. Specifically, we first evaluate the performance on the overall testing set, then compare the results on modality preferred subsets referring to the split strategy. 

As presented in Table~\ref{tab:av_other}, we find some results consistent with the recent works~\cite{peng2022balanced}: all imbalance methods perform better than the unimodal model on the overall set, which proves the effectiveness of multimodal learning. 
Besides, compared with the baseline, all imbalance methods gain significant improvement across different datasets, which indicates that existing methods alleviate the modality imbalance problem to some extent.

Although the imbalance methods achieve great performance on the overall subset, we observe that existing methods consistently perform worse than the unimodal model in the weak-modality preferred subset. 
Concretely, considering the VGG-Sound~\cite{chen2020vggsound} dataset is a curated sound-oriented dataset,  where the audio-preferred subset contains a significantly larger number of samples than the visual-preferred subset. The visual modality is weaker in comparison to the audio modality.
However, Table~\ref{tab:av_other} shows that although the visual unimodal performance is worse than existing imbalance methods on the overall set, the visual unimodal model exceeds other multimodal models with a large margin on the visual-preferred subset, which hints that existing imbalance methods still fail to fully utilize the weak modality.

However, considering the greedy nature of deep learning~\cite{wu2022characterizing}, the multimodal models would inevitably learn the modality biases inherent in these biased audiovisual datasets, which would potentially impact the evaluation of these imbalance methods.
 % Thus, we build a balanced audiovisual dataset to support further analysis of the modality imbalance methods.

% \begin{figure*}[t]
%     \centering
% 		\centering
% 		\includegraphics[width=\textwidth]{materials/datasetStatic.pdf}
%     \caption{Illustrations of our balanced dataset statistics. The histogram demonstrates the amount of each category while the scatter diagram demonstrates the modality discrepancy for each category.}
%     \label{fig:datasetStatic}
% \end{figure*}

\section{The Balanced Audiovisual Dataset}
To alleviate the influence of modality bias and further study the effectiveness of imbalance methods, we build a balanced audiovisual dataset with full consideration of the modality discrepancy. As shown in Figure~\ref{fig:heatmaps}(c), our balanced dataset contains samples with various modality discrepancy, and the allocation of such modality discrepancy is uniformly distributed over the dataset.

To automatically collect videos with corresponding labels, we filter the crawled videos with the off-the-shelf pretrained unimodal models (SlowFast~\cite{fan2020pyslowfast} pretrained on Kinetics-400~\cite{kay2017kinetics} for visual modality and VGGish~\cite{simonyan2014very} pretrained on VGG-Sound~\cite{chen2020vggsound} for audio modality). 
Constrained by the pretrained datasets, we select several categories with the same semantic concept existing in both datasets, then integrate them into 30 categories and crawl videos from YouTube for corresponding topics. 
% (e.g. the ``shaving head'' and ``trimming or shaving beard'' labels in Kinetics-400 are similar to the ``electric shaver, electric razor shaving'' in VGG-Sound), 

Then we extract the full videos into 10-seconds clips and evaluate unimodal confidence for each clip with pretrained model. Referring to the unimodal confidence, we split the clips into three types, called the high-correspondence clips, audio-correct clips, and visual-correct clips to build our dataset, where the high-correspondence clips mean both unimodal models predict true label, audio-correct clips mean only audio modality model predict the true label while the visual-correct clips are on the opposite. With such clip selection strategy, we could preliminarily build a nearly balanced audiovisual dataset, which contains audiovisual clips of three types with similar proportion.

However, the preliminary dataset suffers from long-tailed distribution and slight modality bias due to the limitation of crawling from the website, such as the instruments videos are easy to collect while some human action categories are difficult to obtain, etc. Thus, to deal with such problem, we screen extra samples from VGG-Sound~\cite{chen2020vggsound} and Kinetics-400~\cite{kay2017kinetics} and incorporate them into our dataset. 
In the end, we estimate the modality discrepancy of our dataset and recalibrate by removing noise and some data from dominated modality referring to such estimation.
Finally, our final overall dataset retains 10,000 samples from the Kinetics-400 dataset, 8,000 samples from the VGG-Sound dataset, and 16,000 samples collected from YouTube.
Our dataset remains a uniformly distributed modality discrepancy over the dataset, as shown in Figure~\ref{fig:heatmaps}(c).

\section{Experiments on Our Dataset}

\subsection{Testing on modality-preferred subset}
As Table~\ref{tab:our_different_degree_all} shows, when trained on our balanced audiovisual dataset, all the imbalance methods exhibit superior performance than the baseline and the unimodal model on the overall set, which proves the effectiveness of existing imbalance methods.

However, there still exists no method that exceeds all unimodal models on every modality-preferred subset. The Baseline, OGM-GE~\cite{peng2022balanced} and Grad-Blending~\cite{wang2020makes} methods gain significant improvement on the audio-preferred subset, but fail to perform better than the visual unimodal model in the visual-preferred subset. While the Greedy~\cite{wu2022characterizing} method performs better than the visual unimodal model but less than the audio unimodal model in the audio-preferred subset. 

The results of Table~\ref{tab:our_different_degree_all} indicate that although existing imbalance methods have achieved notable enhancements, they still have difficulty handling samples with modality discrepancy. To analyze the influence of the modality discrepancy on multimodal models, we propose to evaluate these methods with various modality discrepancy subsets. 
% It seems that existing imbalance methods only provide a compromise between modalities, instead fully exploiting multiple modalities to the best.
% Although the results of existing experiments consistently demonstrate such conjecture, it cannot reach a solid conclusion due to the influence of different datasets. Thus, we conduct controlled experiments to verify this conjecture.

\begin{table}[t]
\centering
\resizebox{\linewidth}{!}{
\begin{tabular}{c|ccc}
\toprule
Methods       & Overall     & Audio preferred & Visual preferred \\ \midrule
Baseline      & 70.47    & 78.62        & 59.37           \\
OGM-GE        & \textbf{72.27}    & 78.92     & 63.59           \\
Grad-Blending & 72.04  & \textbf{79.49}          & 61.89    \\
Greedy        & 72.01  & 72.46          & \textbf{71.41}          \\ \midrule
Audio unimodal    & 62.96   & 77.76          & 44.79            \\
Visual unimodal   & 49.82   & 36.79          & 67.58          \\
\bottomrule
\end{tabular}}
    \caption{Evaluation results on our dataset.}
    \label{tab:our_different_degree_all}
    \vspace{-1em}
\end{table}

\subsection{Testing with different modality discrepancy}

Given we have obtained the average unimodal confidence $\hat u_i^m$ of each sample, we propose a threshold $T$ to split the testing subset. Concretely, if $\hat{u_i^a} - \hat{u_i^v} > T$, the sample would be split into the audio-dominated subset, while if $\hat{u_i^v} - \hat{u_i^a} > T$, the sample would be split into the visual-dominated subset. When threshold $T$ increases, the modality discrepancy in the corresponding subset would be larger. In experiments, we set the threshold $T$ as 0, 0.2, and 0.4 to select the subset with different modality discrepancies. 

\begin{figure}[!t]
    \centering
		\includegraphics[width=0.9\linewidth]{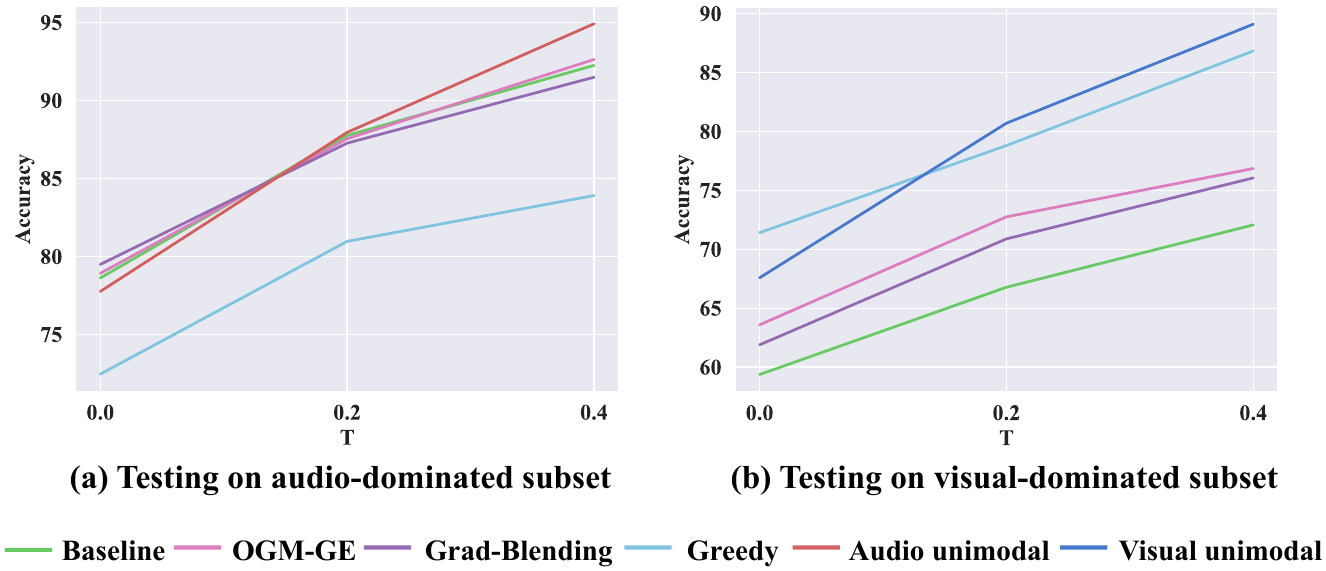}

    \caption{Evaluation results on modality-dominated subsets with different threshold $T$. }
    \label{fig:subset_results}
    \vspace{-1em}
\end{figure}

We provide the results with different thresholds $T$ on the audio-dominated subset and the visual-dominated subset in Figure~\ref{fig:subset_results}. When we only consider the $T=0$ scenarios, there exist some multimodal models exceeding the unimodal model on the corresponding subset (e.g., the OGM-GE in the audio-dominated subset and the Greedy in the visual-dominated subset). However, when the threshold $T$ increases, only the samples with larger modality discrepancy would remain in the modality-dominated subset, and the unimodal model performs better than other multimodal models on the corresponding modality-dominated subset (The audio model performs better than other imbalance methods on the audio dominated subset when $T = 0.4$). 

The multimodal models are expected to perform better than unimodal models, however, 
the extra information instead hinders the performance on multimodal data of large modality discrepancy. Considering the samples with larger modality discrepancy would have more modality noise, these results indicate that the modality noise would affect the performance of multimodal models.

\section{Conclusion}
In this work, we first introduce a modality discrepancy estimation metric and find that existing audiovisual datasets suffer from serious modality bias. To support the comprehensive analysis of the modality imbalance methods, we build a balanced audiovisual dataset with uniformly distributed modality discrepancies. Further, we evaluate existing imbalance methods with different modality discrepancy subsets and find that multimodal models would exhibit inferior performance than the unimodal one in some scenarios with serious modality discrepancy. This comprehensive evaluation method with different modality discrepancies should be considered to make sure the reliability of multimodal models in different application scenarios.

\section{Acknowledgement}
This research was supported by National Natural Science
Foundation of China (NO.62106272), the Young Elite Scientists Sponsorship Program by CAST (2021QNRC001), and
Public Computing Cloud, Renmin University of China.

{\small
\bibliographystyle{ieee}
\bibliography{example_bib}
}

\end{document}